\documentclass[]{mc_nankai}

\usepackage[utf8]{inputenc} 
\usepackage[T1]{fontenc}    
\usepackage{hyperref}       
\usepackage{url}            
\usepackage{booktabs}       
\usepackage{amsfonts}       
\usepackage{nicefrac}       
\usepackage{microtype}      
\usepackage{xcolor}         
\usepackage{array}

\usepackage{amsmath}
\usepackage{url}
\usepackage{graphicx}
\usepackage{multirow}
\usepackage{multicol}
\usepackage{animate}
\usepackage{listings}
\usepackage{algorithm}
\usepackage{algpseudocode}
\usepackage{enumitem}
\usepackage{silence}
\usepackage{pifont}  

\newcommand{\cmark}{\ding{51}} 
\newcommand{\xmark}{\ding{55}} 

\usepackage{wrapfig}  

\newcommand{\nameofmethod}{OneVAE} 

\title{\nameofmethod{}: Joint Discrete and Continuous Optimization Helps Discrete Video VAE Train Better}

\author[1,2]{Yupeng Zhou}
\author[3]{Zhen Li}
\author[1]{Ziheng Ouyang}
\author[1]{Yuming Chen}
\author[2]{Ruoyi Du}
\author[4]{Daquan Zhou}
\author[2]{Bin Fu} 
\author[2]{Yihao Liu} 
\author[2]{Peng Gao}
\author[1]{Ming-Ming Cheng}
\author[1]{Qibin Hou$^{\dagger}$}

\affiliation[1]{VCIP, School of Computer Science, Nankai University}
\affiliation[2]{Shanghai AI Laboratory}
\affiliation[3]{The Chinese University of Hong Kong}
\affiliation[4]{Peking University}

\affiliation[]{$^{\dagger}$Corresponding author}
\vspace{-50pt}
\abstract{
Encoding videos into discrete tokens could align with text tokens to facilitate concise and unified multi-modal LLMs, yet introducing significant spatiotemporal compression compared to continuous video representation.
Previous discrete video VAEs experienced unstable training, long training time, and degraded reconstruction quality.
Given the easier training and superior performance of continuous VAEs, an intuitive idea is to enhance discrete video VAEs by leveraging continuous VAEs.
After rethinking the intrinsic link between discrete and continuous representations, we found that FSQ could effectively preserve pre-trained continuous VAE priors compared to other quantization methods. By leveraging continuous VAE priors, it converges several times faster than training from scratch and achieves superior performance at convergence.
Meanwhile, two structural improvements are proposed. First, inspired by how continuous VAEs enhance reconstruction via enlarged latent dimensions, we introduce a multi-token quantization mechanism, which achieves nearly a 1 dB improvement in PSNR without compromising the token compression ratio.
Second, to tackle reconstruction challenges in high-compression video VAEs, we strengthen first-frame reconstruction, enabling the causal VAE to leverage this information in subsequent frames and markedly improving the performance of $4\times16\times16$ discrete VAEs.
Furthermore, we propose a joint discrete–continuous optimization scheme that unifies the two paradigms and, for the first time, achieves competitive performance on both continuous and discrete representations within a single network. We name our method \nameofmethod{} to reflect this connection.
}
\mclink[Code]{\url{https://github.com/HVision-NKU/OneVAE}}

\begin{document}

\maketitle
\justifying

\section{Introduction}
\label{sec:intro}
\begin{wrapfigure}{r}{0.50\linewidth}  
    \centering
    \vspace{-55pt}  
    \includegraphics[width=0.9\linewidth,height=6cm]{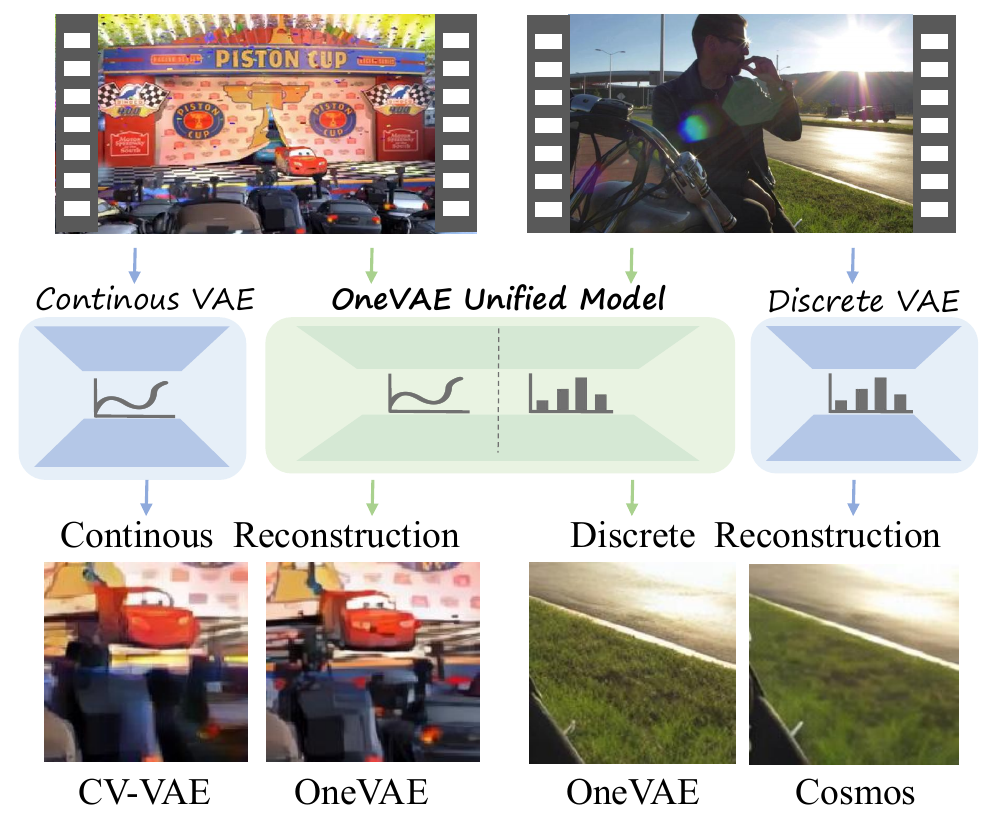}
     \vspace{-10pt}  
    \caption{Illustration of \nameofmethod{}, which could encode both discrete and continuous representations, distinguishing from traditional discrete-only~\cite{cosmos_token} and continuous-only~\cite{zhao2024cv} VAEs. }
    \label{fig:teaser}
    \vspace{-10pt}
\end{wrapfigure}

Auto-regressive language LLMs have shown strong performance, sparking interest in extending their capabilities to image and video generation. In particular, a series of autoregressive generative  models~\cite{wang2024emu3,yu2023language,sun2024autoregressive} have already achieved impressive results in this direction.
A key challenge in extending these models to visual domains is that raw visual data (e.g., RGB pixels) is far more complex than plain text, which raises issues in how to represent such visual content into tokens.
Looking back at diffusion-based generation works, the pioneering work Stable Diffusion~\cite{rombach2022high} succeeded in introducing continuous VAE to enable the generation of photorealistic images with limited computational resources. Since then, the field of AIGC research has advanced rapidly, with a series of generative models~\cite{chen2023pixart,esser2024scaling, podell2023sdxl, yang2024cogvideox, kong2024hunyuanvideo, ma2025step} providing more powerful visual generation results.
However, continuous representations do not align well with LLMs for pure next-token prediction, as it is difficult to directly predict continuous tokens~\cite{li2024autoregressive}.

Previous auto-regress generative models~\cite{sun2024autoregressive,chen2025janus,team2024chameleon} introduced discrete VQ-VAEs~\cite{esser2021taming} to encode visual content, align them with text tokens, and allow a unified autoregressive next-token prediction architecture. 
Despite these impressive results, the VQ-VAEs employed in previous work face two key limitations.
First, due to the heightened difficulty in balancing compression efficiency and information preservation in discrete representations compared to continuous ones, their reconstruction quality still leaves room for enhancement.
Second, as these works have only focused on image generation, their VQ-VAEs are confined to encoding images alone. 

Compared to their image-based counterparts, discrete video VAEs must additionally compress temporal dimension information, rendering the design of a robust discrete video VAE far more demanding.
In the early exploration of discrete video VAE, most works also relied on VQ-based quantization: VideoGPT~\cite{yan2021videogpt} adopted a relatively low compression ratio of $4\times4\times4$ to preserve performance, while OmniTokenizer~\cite{wang2024omnitokenizer} introduced a hybrid image-video encoder using $4\times8\times8$ configuration. EMU-3~\cite{wang2024emu3} further improved the reconstruction quality of the VQ-VAE.
Due to the inherent training instability of VQ, the Cosmo Tokenizer~\cite{cosmos_token} incorporated the quantization of FSQ~\cite{mentzer2023finite} into its VAE architecture. Using FSQ’s improved stability, it was able to support multiple VAEs with varying compression ratios.
Although significantly improving the reconstruction performance of discrete VAEs, the discrete Cosmos Tokenizers still lag behind their continuous counterparts. 

This discrepancy motivates us to further explore ways to enhance the performance of discrete VAEs in this paper.
We first engaged in rethinking the distinctions between continuous and discrete encoding.
For current discrete tokenizers, video sequences are compressed into continuous features, and different quantization methods are employed to quantize these continuous features.
Based on how the features are processed, we categorize the quantization methods into two types: (1) codebook lookup-based ones, which aim to find the vector with the closest distance in the codebook; (2) rounding operation-based ones, which directly round on the mapped features.
Unlike VQ~\cite{esser2021taming}, which uses codebook distance, FSQ~\cite{mentzer2023finite} maps continuous features into a bounded range and rounds them, making it closer to continuous encoding.
We verified this by training a lightweight quantizer in the latent space of continuous VAEs, optimizing only its parameters. As shown in \tabref{tab:vq_vs_fsq}, FSQ outperforms VQ (PSNR 26.46 vs. 22.82).
This confirms FSQ’s closer relation to continuous VAEs and suggests continuous VAEs can support discrete ones, opposing the view that they are independent training~\cite{cosmos_token}.

Based on these observations, we propose a progressive training strategy to accelerate convergence. We start with an $8\times8\times4$ continuous tokenizer, then expand to an $8\times8\times4$ discrete tokenizer and a $16\times16\times4$ continuous tokenizer, and finally transition to a $16\times16\times4$ discrete tokenizer. We choose pretrained VAEs~\cite{kong2024hunyuanvideo,wan2025wan} as the starting point for training, as such models are typically trained on large non-public higher-quality datasets. Our goal is for the progressive pipeline to preserve these useful priors.
As shown in \figref{fig:psnr_training}, unlike methods such as Cosmos Tokenizer~\cite{cosmos_token} that train separate continuous and discrete tokenizers from scratch, our progressive approach achieves competitive performance with significantly fewer training iterations, greatly accelerating training.
Moreover, the experimental results support that our model achieves better reconstruction performance.
Building on this, we further propose an ambitious idea: Could a single model simultaneously learn both discrete and continuous representations?
To explore this, we propose a dual-path joint optimization strategy, which includes an additional FSQ-based discretization branch and a continuous branch, allowing the model to encode both continuous and discrete data during training. 
The experimental results give a clear confirmation. Our unified model can effectively handle both representations, and ablation studies show that it even improves certain metrics, further demonstrating their intrinsic connection and mutual reinforcement.

We also introduce meaningful structural enhancements, including multi-token quantization, first-frame enhancement for causal VAEs, and decoder expansion, to improve performance under high compression ratios. Multi-token quantization alleviates codebook size constraints by encoding each feature position into multiple tokens (two in our experiments). The first frame enhancement stems from our observation that, in causal VAEs, the first frame often suffers from lower quality than subsequent frames due to limited accessible information. By reducing the compression ratio of the first frame only, we significantly improve reconstruction quality for both the first and subsequent frames. Finally, decoder expansion enhance reconstruction capacity by increasing the model’s parameters. Together, these modifications lead to substantial performance gains.
Our contributions are summarized as follows:
\begin{itemize}
\item We propose a progressive training strategy that transitions from a pretrained continuous VAE to discrete VAEs, effectively leveraging prior knowledge to accelerate convergence.
\item We introduce structural enhancements including multi-token quantization and first-frame enhancement for causal VAEs, both of which substantially improve reconstruction quality.
\item We develop a unified tokenizer that jointly models discrete and continuous representations, achieving superior reconstruction performance for both.
\end{itemize}

\section{Relate Work}
\subsection{Generation Model}

Generative models have demonstrated remarkable performance across various domains. In the early stages, Generative Adversarial Networks (GANs)~\cite{goodfellow2014generative, karras2019style, brock2018large} emerged as one of the pioneering and most influential visual generation methods in the deep learning era.
Recently, diffusion models~\cite{ho2020denoising} have demonstrated unprecedented performance in generative tasks, notably in the domain of text-to-image synthesis~\cite{ramesh2022hierarchical, rombach2022high, saharia2022photorealistic}, through a systematic noise removal process.
Subsequent advancements extended diffusion-based methodologies to video generation. Some video diffusion models directly model pixel-value distributions~\cite{li2022efficient, singer2022make}, while others concentrate on modeling the distributions of latent tokens, typically extracted using Variational Autoencoders (VAEs)~\cite{blattmann2023align, wang2023modelscope}, rather than directly handling pixel data.
As an alternative technical approach, autoregressive models~\cite{esser2021taming, ramesh2021zero, yu2022scaling} are a class of generative models widely used to predict the next token in a sequence, inspired by GPT-style~\cite{radford2018improving} architectures.
To reduce computational cost and enhance the quality of generated outputs, these methods rely on discrete or continuous VAEs, leveraging their ability to perform compression-based high-fidelity reconstruction.

\subsection{Discrete VAEs}
Recent advancements in video generation have been significantly involved with the auto-regressive models, driven by the rising popularity of large language models.
This approach requires a discrete tokenizer to quantize visual information by mapping input images into a latent space and identifying the nearest codebook vectors.
Notable models such as TATS~\cite{ge2022long}, MAGVIT~\cite{yu2023magvit}, and VideoGPT~\cite{yan2021videogpt} leverage discrete token training within the VQVAE~\cite{van2017neural} framework, employing 3D VAEs to extract discrete tokens.
TATS~\cite{ge2022long}, for example, introduces a hierarchical sampling strategy with autoregressive and interpolation Transformers, improving both generation quality and efficiency.
MAGVIT-v2~\cite{yu2023language} refines the VQ-VAE codebook by reducing its embedding dimension to zero and incorporating Lookup-Free Quantization (LFQ), which has been widely adopted in contemporary models. 
VideoGPT~\cite{yan2021videogpt} utilizes VQ-VAE, applying 3D convolutions and axial self-attention to learn downsampled discrete latent representations of video data.
Additionally, newer models, such as LaViT~\cite{jin2024video} and ElasticTok~\cite{yan2024elastictok}, generate dynamic discrete visual tokens while preserving high-level semantic information. 
Approaches like the Omni-tokenizer~\cite{wang2024omnitokenizer}, a joint image and video tokenizer, 
and Cosmos-Tokenizer~\cite{cosmos_token}, which uses Finite Scalar Quantization, further enhance the discrete tokenization process. 
At the same time, different improvements for discretization methods have been proposed in combination with diverse model architectures, such as VQ~\cite{esser2021taming}, LFQ~\cite{yu2023language}, CVQ~\cite{zheng2023online}, IBQ~\cite{shi2024taming}, and FSQ~\cite{mentzer2023finite}. These strategies optimize the tokenization process, significantly contributing to advancements in visual generation.
Among these studies, 
CVQ alleviates code collapse by learning the cluster codebook.
LFQ improves upon the original VQ by eliminating the lookup step.
IBQ improves the discretization process by modifying the gradient calculation method for optimizing the codebook training.
and FSQ simplifies quantization by using a fixed mapping and rounding operation, improving stability.

\subsection{Continuous VAE}

Compared to discrete tokenization, continuous tokenization~\cite{zhao2024cv, chen2024od, tang2024vidtok, yang2024cogvideox} typically offers higher reconstruction fidelity. 
In the field of image generation, Stable Diffusion~\cite{rombach2022high} introduced the use of continuous VAEs in diffusion models, significantly enhancing image generation. This approach has inspired numerous subsequent works~\cite{chen2023pixart, esser2024scaling, podell2023sdxl} in the field of video generation.
Some studies integrate 3D convolutions or spatio-temporal attention mechanisms into the backbone network~\cite{bar2024lumiere, ho2022video, singer2022make, wu2023tune, jiang2024dive}, creating latent spaces specifically designed for video data. 
Video generation models initially performed frame-by-frame decoding, but following Sora~\cite{videoworldsimulators2024}, a series of video diffusion models trained video VAEs to compress data along the temporal dimension~\cite{kong2024hunyuanvideo,li2024wf,opensora,pku_yuan_lab,yang2024cogvideox}. 
Some works like Hunyuan Video~\cite{kong2024hunyuanvideo} and Allegro VAE~\cite{li2024wf} also contributed to this paradigm shift.
Open-Sora~\cite{opensora} and Open-SoraPlan ~\cite{pku_yuan_lab} are open-source projects aimed at replicating OpenAI’s Sora, offering efficient continuous VAEs. 
Meanwhile, CogVideoX~\cite{yang2024cogvideox} retains more information by preserving a larger number of latent channels. 
Additionally, Cosmos-Tokenizer~\cite{cosmos_token} provides a suite of continuous video tokenizers with various compression ratios. 
CV-VAE~\cite{zhao2024cv} employs latent space regularization to align the video latent space with the latent space of existing image VAEs~\cite{rombach2022high}, enabling effective spatio-temporal compression. 
In contrast to previous approaches that handle discrete and continuous representations in isolation, we introduce \nameofmethod{}, which unifies both types within a single framework.
By combining progressive training strategies and architectural modifications, our method achieves improved reconstruction quality for discrete representations while maintaining strong performance across representation types.

\section{Method}
\subsection{Preliminaries of Continuous and Discrete VAEs}
\label{sec:preliminary}
Continuous and discrete VAEs differ fundamentally in how they represent latent variables. Continuous VAEs~\cite{kingma2013auto} model the latent space as a continuous probability distribution, where the encoder \( E(x) \) maps the input $x$ to the parameters of a distribution—typically a Gaussian. The latent variable $z$ is then sampled from this distribution.

\begin{equation}
E(x) = \mathcal{N}(\mu, \sigma^2), \quad z \sim E(x)
\end{equation}
where \( \mathcal{N}(\mu, \sigma^2) \) represents a Gaussian distribution with mean \( \mu \) and variance \( \sigma^2 \).

Same as continuous VAEs, discrete VAEs compress video data $V \in \mathbb{R}^{C\times T \times H\times W}$ into a low-resolution representation $z \in \mathbb{R}^{D \times \frac{T}{t} \times \frac{H}{h} \times \frac{W}{w}}$ through the encoder and then decompress back to the original representation  $\hat{x}$ using the decoder:
\begin{equation}
    z = E(x), \quad   \hat{x} = D(z)
\end{equation}
where \( E(\cdot) \) represents the encoder and \( D(\cdot) \) is the decoder.  Discrete VAEs~\cite{shi2024taming,esser2021taming} typically remove the Gaussian sampling process used in continuous VAEs. To convert $ z $  into a discrete token, two primary quantization strategies exist. VQ-based quantization selects the closest vector from a predefined codebook 
\( \mathcal{C} = \{ c_i \}_{i=1}^{N} \) based on Euclidean distance:
\begin{equation}
    q_i = \arg\min_{c_i \in \mathcal{C}} \| z_i - c_i \|_2.
\end{equation}
Alternatively, FSQ-based quantization~\cite{mentzer2023finite} directly constrains each dimension of 
$z$ within a bounded range and applies a rounding operation:
\begin{equation}
q_i = \mathrm{Round}( \mathrm{Bound}(z_i) ).
\end{equation}
After that, the $d$ dimensions of $q_i$ are all integers in $(-L/2, L/2)$, and they can be easily converted into an index.

\begin{wrapfigure}{r}{0.5\textwidth} 
    \centering
    \small
    \begin{tabular}{>{\raggedright\arraybackslash}p{2.7cm}ccc}
        \toprule
        \textbf{Quant. Method} & \textbf{PSNR} $\uparrow$ & \textbf{SSIM} $\uparrow$ & \textbf{LPIPS} $\downarrow$ \\
        \midrule
        FSQ  & 26.46 & 0.8566 & 0.0825 \\
        VQ   & 22.82 & 0.7310 & 0.2071 \\
        \bottomrule
    \end{tabular}
    \caption{Comparison of VQ~\cite{esser2021taming} and FSQ~\cite{mentzer2023finite} trained in the latent space of a continuous VAE, where only the intermediate quantization modules are trained.}
    \label{tab:vq_vs_fsq}
\end{wrapfigure}
Intuitively, the FSQ-based approach involves only a simple mapping and rounding of continuous features, which makes it highly similar to those features. To explore this, we experiment in \tabref{tab:vq_vs_fsq} by training quantizer modules based on FSQ and VQ into a pre-trained continuous VAE to train using discrete reconstruction loss. The results show that FSQ outperforms VQ, suggesting that the discrete features encoded by FSQ exhibit a higher similarity to the continuous features of the VAE.

\begin{figure*}[!tp]
    \centering
    \includegraphics[width=\linewidth]{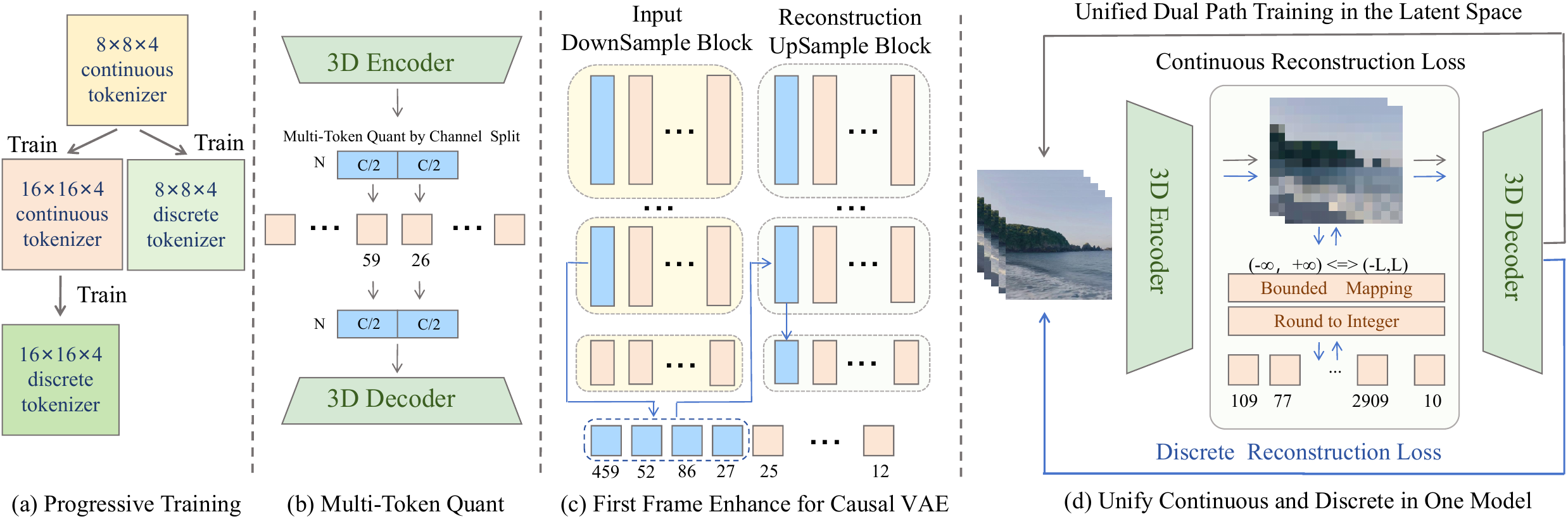}
    \caption{Training method and architecture of the proposed model. Figure (a) illustrates the progressive training process, where all models are derived from an 8×8×4 continuous VAE, which can be either the HunyuanVideo VAE~\cite{kong2024hunyuanvideo} or the Wan VAE~\cite{wan2025wan} to provide pre-trained data priors. Figure (b) demonstrates the working principle of multi-token quantization, which can significantly enhance the representational capacity. Figure (c) illustrates the First-Frame Enhancement mechanism: by reducing the compression strength of the first frame to strengthen its reconstruction, it enables the causal VAE to leverage this information in subsequent frames, markedly improving the 4×16×16 discrete encoder’s reconstruction quality.  Figure (d) presents the unified training framework that integrates both continuous and discrete paths in the latent space. In the continuous path, the 3D encoder maps the input data to a continuous latent representation (C×T×H×W), which is optimized by the continuous reconstruction loss. In the discrete path, after bounded mapping and integer rounding, the FSQ quantization method converts the continuous latent representation into discrete tokens (C×T×H×W→N Token), which are trained with the discrete reconstruction loss. This approach enables the model to handle continuous and discrete data simultaneously with better performance for both.}
    \label{fig:pipeline}
\end{figure*}
\subsection{Improved Training Procedure and Structure}
\label{sec:improved_training}

\myPara{Tree-Structured Progressive Training.} 
VAEs are traditionally trained separately. To accommodate different compression ratios, both discrete and continuous VAEs may need to be trained as a suite of multiple models, leading to significant computational overhead. Moreover, as shown in the training details of Open-SORA~\cite{opensora} and Open-SORA-Plan~\cite{lin2024open}, video VAEs often require extended training times.
Additionally, compared to continuous VAEs, discrete VAEs suffer significant information loss due to the quantization process and are prone to issues such as code collapse~\cite{esser2021taming}, making training increasingly challenging.
Based on the analysis presented in \secref{sec:preliminary} and the experimental results in \tabref{tab:vq_vs_fsq}, we argue that continuous VAEs can serve as an effective warm-up for training discrete VAEs. The key insight is to leverage the latent space representations learned by the continuous VAE to facilitate the training of the discrete VAE.
This approach allows the model to start training from a well-optimized state rather than from random initializations.
In our training pattern, we first train a 8$\times$8$\times$4 continuous VAE and then gradually extend it to both discrete and continuous VAEs with 8$\times$8$\times$4 and 16$\times$16$\times$4 configurations.
For converting the continuous VAE to a discrete VAE, we simply insert an FSQ quantization module into the middle of the pre-trained continuous VAE to discretize it into tokens.
Afterward, we fine-tune the model to optimize it for discrete representations.
To further minimize discrepancies, we initialize the discrete VAE from a continuous VAE with the same downsampling ratio.
Therefore, for high-compression discrete VAEs, we first obtain a high-compression continuous VAE through progressive training and then convert it into a discrete VAE using the same approach.
As shown in \figref{fig:pipeline}(a), we refer to this approach as tree-structured progressive training, where all models are derived from a 8$\times$8$\times$4 continuous VAE. By adopting this progressive training strategy, we ensure that the training process remains manageable, allowing the model to learn effectively at each stage before advancing to the next level of complexity.

\myPara{Multi-Token Quantization} 
As illustrated in \figref{fig:pipeline} (b), we propose a multi-token quantization strategy that decomposes each latent feature vector along its channel dimension into multiple smaller sub-vectors, and quantizes each independently. Specifically, each feature channel is divided into several segments—for example, two equal parts in our implementation. Each segment is then quantized separately into discrete tokens using shared codebooks.
This approach increases the effective representational capacity without enlarging the total codebook size, as multiple tokens collectively encode the original vector with finer granularity. By capturing more subtle variations within each channel, multi-token quantization enables a richer and more flexible discrete representation of latent features. As a result, this leads to improved reconstruction fidelity and overall performance.
Compared to conventional single-token quantization that compresses an entire channel vector into one token, our method strikes a balance between compression and expressiveness, addressing the limited expressiveness of traditional vector quantization methods under high compression ratios. This design constitutes a key contribution of our work, demonstrating substantial performance improvement.

\begin{wrapfigure}{r}{0.50\textwidth} 
    \centering
            \vspace{-15pt}
    \includegraphics[width=0.50\textwidth]{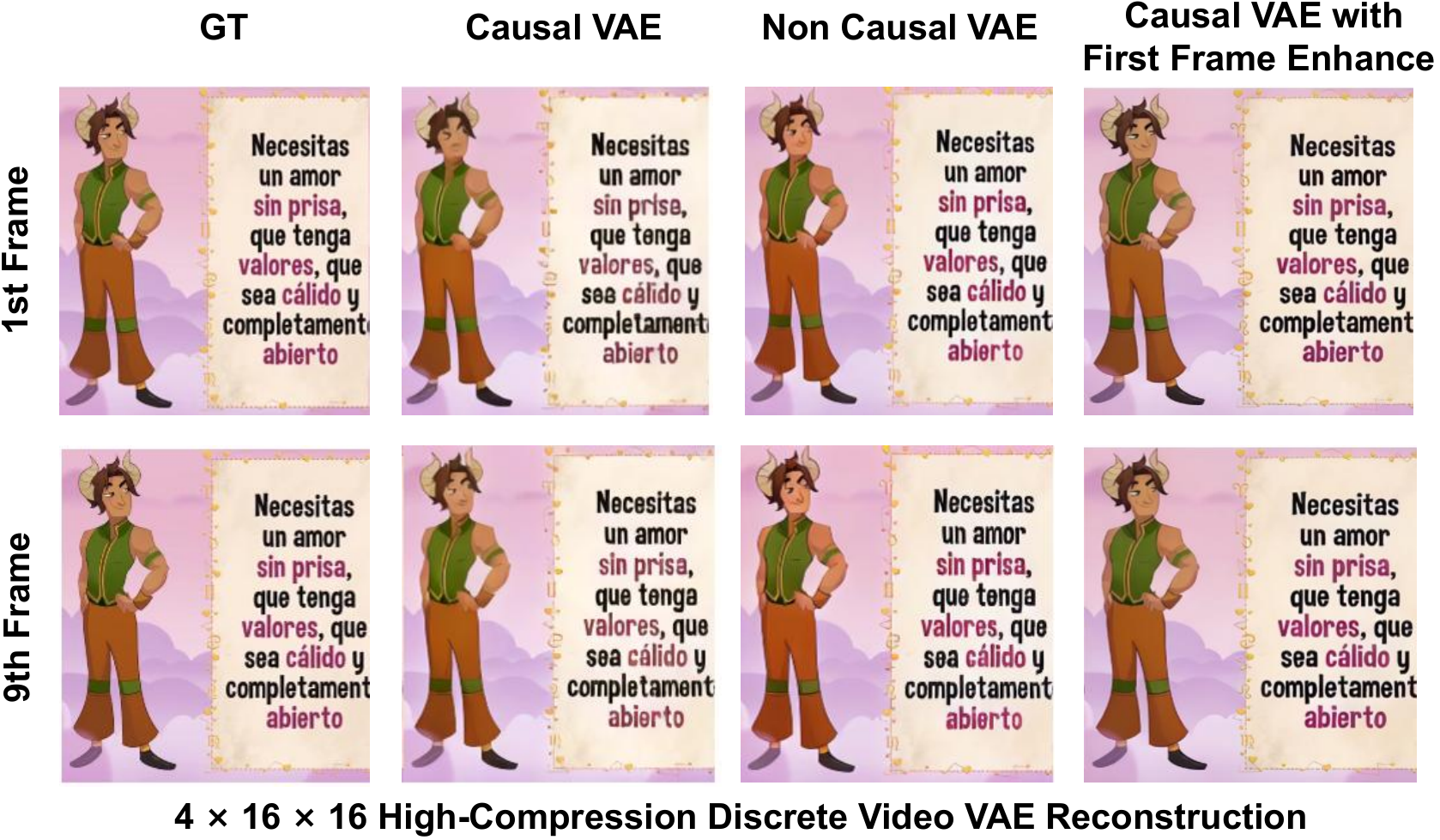}
    \vspace{-10pt}
    \caption{Visual comparison illustrating our First-Frame Enhancement strategy. The second column shows results from a causal $4\times16\times16$  video VAE, where the first frame suffers from noticeably poorer reconstruction quality than subsequent frames, and the overall performance lags behind the non-causal VAE in the third column. With our First-Frame Enhancement, the fourth column shows substantial improvements in both the first and subsequent frames. Zoom in to examine text reconstruction details.}
    \label{fig:compare_noncausal}
\end{wrapfigure} 
\myPara{First-Frame Enhancement}
Previous works on causal VAEs~\cite{cosmos_token,wang2024loong,tang2024vidtok} have either explicitly noted or been implicitly observed (e.g., in open-source model outputs) that high compression video VAE tend to degrade the quality of early frames in a sequence. However, this issue has not been effectively addressed. By comparing the same VAE architecture under causal and non-causal settings in \figref{fig:compare_noncausal}, we found that the inferior quality of early frames in the causal setting likely stems from their limited accessible information.
As illustrated in \figref{fig:pipeline}(c), we propose a \emph{First-Frame Enhancement} strategy to address this issue. The key idea is to reduce the compression strength of the first frame by allocating more tokens per spatial unit, while keeping the overall video compression ratio unchanged.
This design enriches the structural and textural fidelity of the anchor frame, which in turn serves as a higher-quality reference for reconstructing subsequent frames in a causal manner.
As shown in \figref{fig:compare_noncausal}, compared to uniform high-ratio compression, which can severely degrade early-frame details and disrupt temporal consistency, our method preserves anchor-frame fidelity without increasing the token budget. This targeted adjustment significantly improves both the first-frame reconstruction and the temporal coherence of later frames, leading to substantial performance gains in high-compression discrete VAE.

\myPara{Expanding Decoder.} 
In high-compression discrete VAEs, quantization leads to a significant reduction in the amount of information retained in the latent space, which directly impacts the quality of the output.
Therefore, the original decoder faces a challenge, as it must compensate for this loss while generating meaningful and coherent results, which is responsible for reconstructing the data.
We argue that the reconstruction module needs to be larger to effectively handle the associated complexities. 
When training the 16$\times$16$\times$4 discrete VAE, we insert additional layers into the decoder to improve reconstruction quality.
%
As shown in \tabref{tab:ablation}, by scaling the decoder, we provide it with more expressive power, enabling it to better reconstruct the data despite the challenges posed by discretization, which leads to a noticeable improvement in reconstruction performance.

\subsection{Joint Optimization For Unify VAE}
\label{sec:joint}

\myPara{Joint Discrete and Continuous Dual Path Training.}
Building on the improved training techniques outlined in \secref{sec:improved_training}, we introduce our \nameofmethod{}, which adopts a joint discrete and continuous optimization approach to help discrete VAE training. 
As show in \figref{fig:pipeline}(b), our method simultaneously optimizes two pathways during training: One path performs discrete reconstruction, while the other performs continuous reconstruction. 
Specifically, at each training step, we randomly choose whether the latent variable will be encoded discretely or continuously, and the model adjusts its encoding accordingly.
The model is jointly optimized using the loss from discrete reconstruction $L_{d}$ and the loss from continuous reconstruction $L_{c}$. The loss function of our joint discrete and continuous optimization is defined as follows:
\begin{equation}
    L=\left\{\begin{array}{ll}
L_{d}, & \text { if } r< R_{dis} \\
L_{c}, & \text { if } r \geq R_{dis}
\end{array}\right.
\end{equation}
where $R_{dis}$  is a ratio that controls the proportion between discrete and continuous reconstruction, and  $r$ is a random number in the range $(0, 1)$.
We propose this approach for two main reasons. First, to preserve the continuous VAE prior, we introduce a tree-structured progressive training as outlined in \secref{sec:improved_training}, where we initialize a discrete VAE from a continuous VAE as a strong starting point. This joint optimization prevents the disruption of the learned continuous priors during the transition to a discrete representation, ensuring that the underlying structure learned from the continuous VAE continues to guide the training process of the discrete VAE.
The results of the ablation experiments in \tabref{tab:ablation} further validate our conclusions. Through progressive continuous initialization and the continuous-discrete joint optimization, the performance of our model can be improved compared to the baseline.

\myPara{Unified Model.}
Through the explorations above, we discover the interplay between discrete and continuous representations and find that the continuous VAE can significantly aid in training the discrete VAE. 
In our original dual-path training, we introduce a ratio $R_{dis}$ that controls the proportion between discrete and continuous reconstruction loss.
A natural idea that follows is whether this dual-path joint optimization, which simultaneously incorporates both the continuous and discrete paths, can be extended to a unified model capable of simultaneously encoding both continuous and discrete representations.
Since the original dual-path training is primarily designed to enhance the discrete representation, 
$R_{dis}$ is set to a high value to ensure that the discrete path occupies a large proportion.
To explore the unified model, we set the ratio to 0.5 to balance the proportion, which could enable a trade-off between discrete and continuous representations, and train the unified model accordingly.
Surprisingly, we find that with balanced dual patch training, we obtain a unified model that performs well in both discrete and continuous encoding. 
Moreover, as shown in \tabref{tab:baseline}, our unified model achieves strong performance in both discrete and continuous representation tasks. 
The ability to simultaneously handle both types of representations allows the model to leverage their respective strengths, outperforming previous models in both discrete and continuous reconstruction tasks.
Our unified model further proves that discrete and continuous representations are not isolated, breaking the previous notion of training them separately and providing interesting insights for future VAE research.
\section{Experiments}

\subsection{Implementation Details}
\label{sec:imple_details}
\myPara{Training.} 
Following previous work~\cite{zhao2024cv}, we train our model on the WebVid-10M dataset.
AdamW~\cite{loshchilov2017decoupled} optimizer is used in the training, with hyperparameters set to $\beta_1 = 0.5$ and $\beta_2 = 0.9$, following VQGAN~\cite{shi2024taming}. 
To reduce memory overhead, we train the model using 16-bit floating-point precision (FP16).
Before training, video data was preprocessed into 17-frame clips, with each frame resized to 256$\times$256 resolution. We employed a batch size of 8 for all training iterations. 
We also incorporated Temporal PatchGAN~\cite{chang2019free} to enhance the model's generative capabilities.
The GAN training was initiated after a 20K-step warm-up phase, ensuring stable adversarial learning. 

\myPara{Evaluation.}
Follow pervious work~\cite{lin2024open,zhao2024cv}, evaluation is conducted on Panda 70M~\cite{chen2024panda} at a resolution of 33×512×512. We first resize the shorter side of the original video to 512, followed by a central crop.
\begin{wrapfigure}{r}{0.50\textwidth} 
    \centering
    \includegraphics[width=0.50\textwidth]{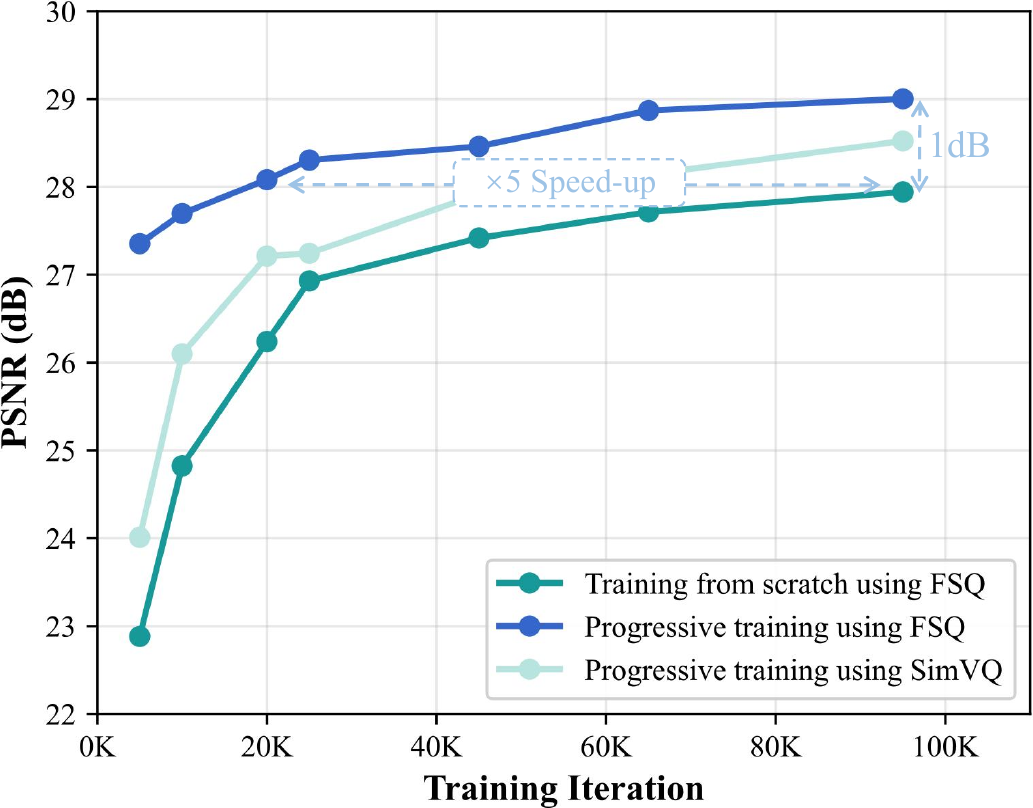}
    \vspace{-15pt}
    \caption{PSNR in training iterations for three different methods. Progressive training with FSQ achieves a significant 5$\times$ convergence speedup compared to training from scratch, and attains better upper-bound performance than both progressive training with SimVQ and training FSQ from scratch.}
    \label{fig:psnr_training}
    \vspace{-40pt}
\end{wrapfigure} 
To evaluate our model, we use multiple metrics across different aspects. 
For reconstruction quality, we employ PSNR, SSIM~\cite{wang2004image}, and LPIPS~\cite{zhang2018unreasonable}. These metrics measure pixel-wise accuracy, structural similarity, and perceptual similarity, respectively.
Furthermore, FVD~\cite{unterthiner2019fvd} is used to assess visual quality and temporal consistency.

\subsection{Ablation Study}

\myPara{Ablation of Progressive Training.}
As shown in \figref{fig:psnr_training}, to verify the effectiveness of our 
progressive training strategy, we compare three training settings: (1) training from scratch using FSQ~\cite{mentzer2023finite}, (2) progressive training with SimVQ~\cite{zhu2024addressing}, and (3) progressive training with FSQ. 
As shown in \figref{fig:psnr_training}, progressive training using FSQ leads to both faster convergence and better final reconstruction quality.
Specifically, the progressive-FSQ strategy achieves a significant 5× speed-up in convergence compared to training FSQ from scratch, reducing the number of iterations needed to reach high-quality results. 
Moreover, it surpasses both baselines in final performance, attaining a PSNR improvement of approximately 1dB. 
Notably, progressive training with SimVQ also accelerates early convergence but saturates at a lower PSNR. 
We attribute the superior final performance of FSQ to its closer alignment with continuous representations, which helps retain useful priors during the transition from continuous to discrete modeling.

\begin{figure}
    \centering
    \includegraphics[width=\linewidth]{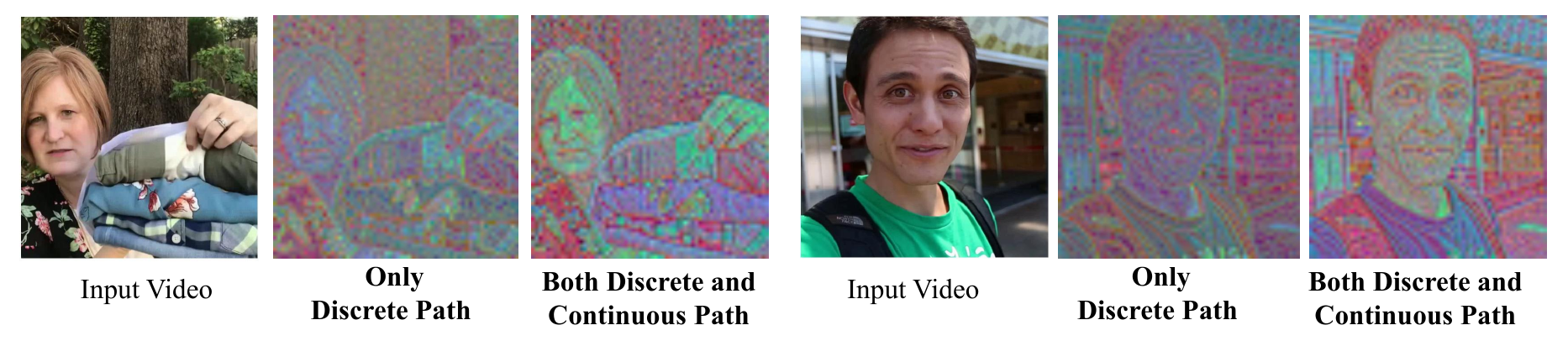}

    \caption{Visual comparison between training with only discrete path and both paths. We visualize and compare the features before quantization. Our proposed joint discrete and continuous paths training enables the model to capture richer features in the latent space for more effective video encoding.}
    \label{fig:vis_latent}
\end{figure}

\myPara{Latent Space Visualization.} To demonstrate the effectiveness of our dual-path optimization, we visualize and compare the features before quantization between the model optimized using only the discrete path and the model optimized using both the continuous and discrete paths. As shown in the \figref{fig:vis_latent}, training with both discrete and continuous paths enables the model to capture richer features in the latent space. We hypothesize that this improvement is due to the enhanced connection between the decoder and encoder in dual-path reconstruction. In discrete tokenizers, backpropagation relies on using the gradients of discrete vectors as approximations for continuous features by stopping gradients, which may introduce inaccuracies in gradient propagation. 
In contrast, our unified training approach likely helps the model achieve more precise gradient flow, leading to more effective video encoding.

\myPara{Expanding the Decoder.}
As mentioned in \secref{sec:improved_training}, to achieve better reconstruction 
\begin{wrapfigure}{r}{0.5\textwidth} 
    \centering
    \small
    \begin{tabular}{c|l|c|c}
        \toprule
        \textbf{Comp. Rate} & \textbf{Method} & \textbf{PSNR} & \textbf{SSIM} \\
        \midrule
        $4\times16\times16$ & Baseline            & 24.86 & 0.7950 \\
        $4\times16\times16$ & + Expanding Decoder & 25.05 & 0.8003 \\
        \bottomrule
    \end{tabular}
    \caption{Expanding the decoder further improves the performance of $4\times16\times16$ compression VAE.}
    \label{tab:ablation}
\end{wrapfigure}
in high-compression (16$\times$16$\times$4) discrete VAEs, we increased the decoder’s capacity by adding additional layers. As shown in \tabref{tab:ablation}, we observed a noticeable improvement in PSNR and SSIM. The increased decoder approach resulted in a PSNR of 25.05 and an SSIM of 0.8003, compared to the baseline performance of 24.86 and 0.7950, respectively. These results suggest that expanding the decoder helps mitigate information loss at high compression ratios and improves overall performance.

\begin{table*}
\centering
\setlength{\tabcolsep}{1mm}
{
\small
\begin{tabular}{lcccccccccc}
\toprule
Method & Compression & $z\_dim$ & Unified Model & Token Type & PSNR ($\uparrow$) & SSIM ($\uparrow$) & LPIPS ($\downarrow$) & FVD ($\downarrow$)  \\
\midrule

CV-VAE & \multirow{4}*{8$\times$8$\times$4} & 4 & \xmark  & \multirow{4}*{Continuous} & 31.12 & 0.9142 & 0.0554 & 100.72  \\
Open-Sora-Plan-v1.2 &  & 4 & \xmark  &  & 27.88 & 0.8840 & 0.0641 &124.66 \\
Open-Sora-v1.2 &  & 4  & \xmark &  & 31.67 & 0.9147 &  0.0648 &  104.48 \\
\textbf{\nameofmethod{}} &  & 6  & \cmark &  & \textbf{32.48} & \textbf{0.9335} &   \textbf{0.0430} & \textbf{59.16} \\
\midrule
Cosmos Tokenizer & \multirow{1}*{16$\times$16$\times$4}  & \multirow{3}*{16}   & \xmark & \multirow{3}*{Continuous} & 30.61 & 0.9144 & 0.0739 & 135.68 \\
Cosmos Tokenizer & \multirow{1}*{16$\times$16$\times$8}  &    & \xmark &  & 29.98 & 0.9041 & 0.0868 & 154.34 \\
\textbf{\nameofmethod{}} &  \multirow{1}*{16$\times$16$\times$4} &  & \cmark &  & \textbf{32.43} & \textbf{0.9318} & \textbf{0.0535} & \textbf{87.61} \\
\midrule
OmniTokenizer & \multirow{4}*{8$\times$8$\times$4} & \multirow{4}*{-} &  \xmark & \multirow{5}*{Discrete} & 23.73 & 0.8974 & 0.0735 & 206.18 \\
Emu-3         & &  &\xmark &  & 29.45 & 0.8912 & 0.0584 & 225.46 \\
Cosmos Tokenizer & &  & \xmark &  & 30.34 & 0.9159 & 0.0524 & 82.86 \\
\textbf{\nameofmethod{}} &  &  & \cmark &  & 30.67 & 0.9228 & 0.0528 & 108.93 \\
\textbf{\nameofmethod{}-MT} & {8$\times$8$\times$8}  &  & \xmark &  & \textbf{31.80} & \textbf{0.9399} & \textbf{0.0450} & \textbf{78.35} \\
\midrule
Cosmos Tokenizer & 8$\times$8$\times$8  & \multirow{3}*{-} & \xmark & \multirow{4}*{Discrete} & 27.17 & 0.8638 & 0.1409 & 303.37 \\
Cosmos Tokenizer & {16$\times$16$\times$4}  &  & \xmark &  & 26.80 & 0.8531 & 0.1260 & 350.87 \\
Cosmos Tokenizer & {16$\times$16$\times$8}  &  & \xmark &  & 26.18 & 0.8342 & 0.1337 & 396.97 \\
\textbf{\nameofmethod{}} & 16$\times$16$\times$4  &  & \cmark &  & 27.40 & 0.8690 & 0.1063 & 336.48 \\
\textbf{\nameofmethod{}-FE} & 16$\times$16$\times$4  &  & \xmark &  & \textbf{28.11} & \textbf{0.8772} & \textbf{0.0814} & \textbf{213.04} \\
\bottomrule
\end{tabular}
}
\caption{Quantitative comparisons of reconstruction quality with baselines on the Panda70M test set demonstrate that our approach outperforms previous tokenizers. \nameofmethod{}-MT and \nameofmethod{}-FE correspond to our proposed multi-token quantization and first-frame enhancement strategies, respectively, as detailed in \secref{sec:improved_training}. \nameofmethod{}-MT employs an 8×8×8 compression ratio, as it quantizes each spatial position into two tokens. }
\label{tab:baseline}
\end{table*}

\subsection{Quantitative Comparison}
As shown in \tabref{tab:baseline}, we provide a quantitative comparison of our proposed unified model, \nameofmethod{}  with several continue-only and discrete-only baseline models, including CV-VAE~\cite{zhao2024cv}, Open-Sora~\cite{opensora}, Open-Sora-Plan~\cite{lin2024open}, OmniTokenizer\cite{wang2024omnitokenizer}, Cosmos Tokenizer~\cite{cosmos_token} and Emu-3~\cite{wang2024emu3}.  Due to the 16$\times$16$\times$4 ratio, lack of comparison models, we modified the Cosmos Tokenizer from 16$\times$16$\times$8. The models are evaluated using various metrics, including PSNR, SSIM, LPIPS, and FVD, which are commonly used to assess the quality of reconstructions and generative performance. From the table, we can observe that \nameofmethod{} consistently outperforms other discrete-only and continuous-only models in most metrics. In summary, the quantitative results demonstrate that our \nameofmethod{} not only unifies discrete and continuous representations but also achieves superior overall reconstruction performance, further corroborating that discrete and continuous representations can be unified.

\begin{figure}[!htp]
    \centering
    \includegraphics[width=\linewidth]{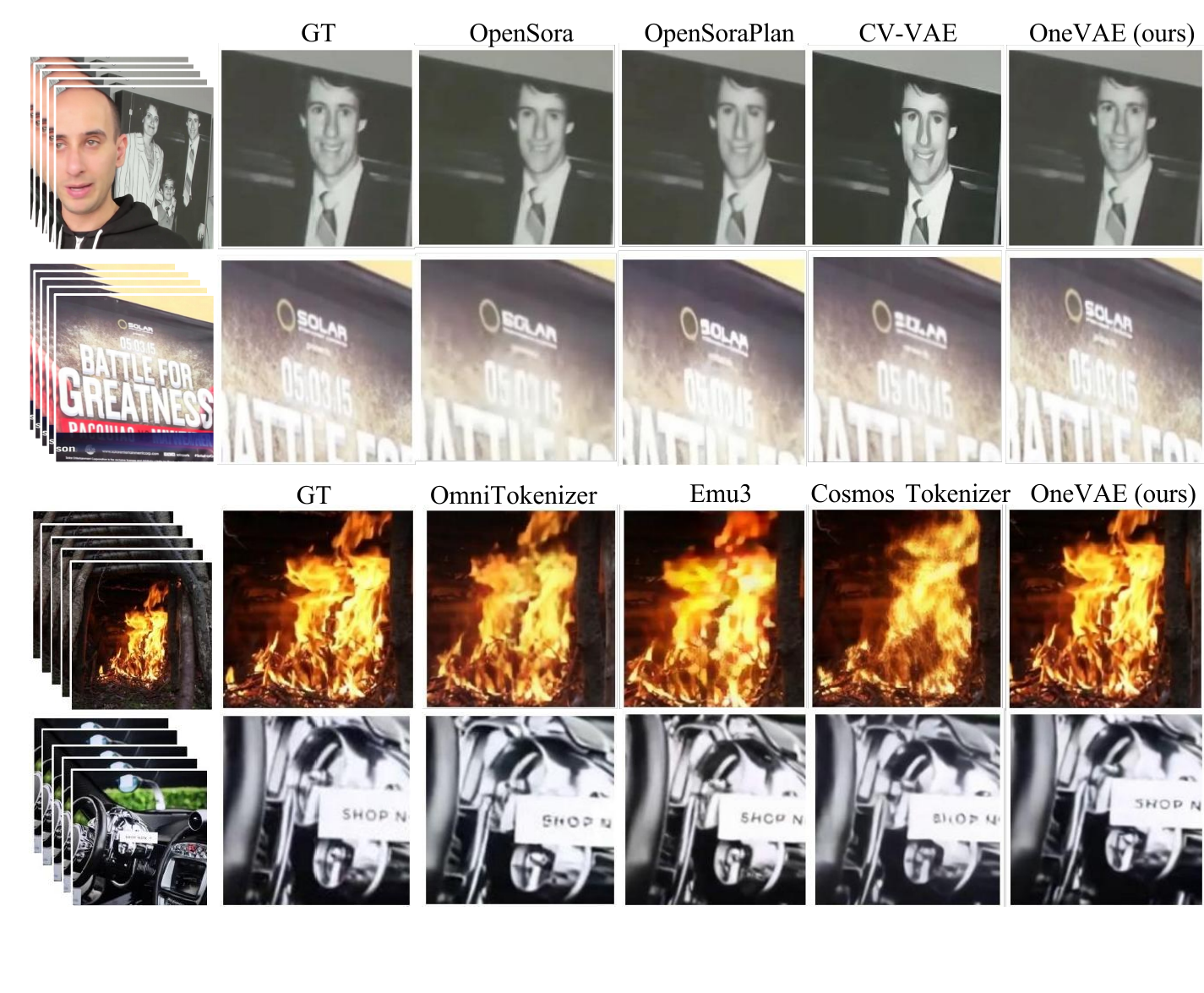}
    \vspace{-35pt}
    \caption{We present a qualitative comparison of our proposed unified model OneVAE with continuous only and discrete only methods, including CV-VAE~\cite{zhao2024cv}, Open-Sora~\cite{opensora}, Open-Sora-Plan~\cite{lin2024open}, OmniTokenizer~\cite{wang2024omnitokenizer}, and Emu-3~\cite{wang2024emu3}.}
    \label{fig:visual_results}
\end{figure}

\subsection{Qualitative Comparison}
As shown in \figref{fig:visual_results}, we present a qualitative comparison 
of the performance of our proposed unified VAE (\nameofmethod{}) against several discrete-only and continuous-only baseline models under the closest model settings.
When compared to continuous-only methods, it is evident that OneVAE produces the most accurate and high-fidelity reconstructions among all models. In the first example, our method reconstructs a clearer and more natural human face. In the second example, OneVAE successfully restores fine text details, demonstrating its superior capability in preserving structural integrity and texture clarity.
When compared to discrete-only tokenizers, OneVAE again stands out by producing more realistic and visually appealing reconstructions. For the first video, \nameofmethod{} reconstructs richer and more detailed flames, whereas other methods produce blurry textures, especially struggling to preserve fine details. In the second comparison, OneVAE reconstructs sharper text, making it the only method among all comparisons that renders the text fully readable.
Overall, despite as a unified discrete and continuous VAE, the qualitative results highlight that OneVAE reconstruct the most accurate and visually coherent videos, demonstrating its superior performance in preserving details and generating high-quality outputs.

\section{Conclusions}
In this work, we propose \nameofmethod{}, a unified VAE framework that bridges discrete and continuous video representations. By rethinking their connection, we propose a progressive training strategy leveraging pretrained continuous VAE priors, which accelerates convergence and improves performance. Furthermore, we propose a unified VAE that performs well on both discrete and continuous representations.
To further enhance video VAE performance, we propose meaningful structural improvements: multi-token quantization to increase representational capacity, and first-frame enhancement to preserve anchor-frame fidelity in causal VAEs and improve temporal consistency. These designs significantly improve reconstruction quality under high compression.
Extensive experiments validate that our approach achieves faster training, better reconstruction quality, and unified modeling capability—offering a promising direction for future video generation and video VAE.

{
\small
\bibliographystyle{plain}
 \bibliography{main}
}

\end{document}